\documentclass[runningheads]{llncs}
\usepackage{graphicx}
\usepackage{tikz}
\usepackage{comment}
\usepackage{amsmath,amssymb} 
\usepackage{color}
\usepackage{multirow}
\usepackage{dsfont,bbm}
\usepackage{hyperref}
\newcommand{\eat}[1]{}
\newcommand{\zyang}[1]{{\color[rgb]{0,0,1}{\tiny\textbf{ZY:}}{\normalsize\itshape#1}}}

\hypersetup{
    colorlinks=true,
    linkcolor=red,
    filecolor=magenta,      
    urlcolor=blue,
}

\def\eg{\emph{e.g}.} 
\def\ie{\emph{i.e}.} 
\def\cf{\emph{cf}. } 
\def\etc{\emph{etc}.} 
 
\def\etal{\emph{et al}.}

\begin{document}
\pagestyle{headings}
\mainmatter
\def\ECCVSubNumber{2152}  

\title{Improving One-stage Visual Grounding by Recursive Sub-query Construction } 

\titlerunning{Improving One-stage VG by Recursive Sub-query Construction} 
\authorrunning{Zhengyuan Yang, Tianlang Chen, Liwei Wang, Jiebo Luo} 
\author{Zhengyuan~Yang$^1$ \quad
Tianlang~Chen$^1$ \quad
Liwei~Wang$^2$ \quad
Jiebo~Luo$^1$
}
\institute{$^1$University of Rochester \quad $^2$Tencent AI Lab, Bellevue\\
\email{\{zyang39,tchen45,jluo\}@cs.rochester.edu, liweiwang@tencent.com}
}

\maketitle

\begin{abstract}
  We improve one-stage visual grounding by addressing current limitations on grounding long and complex queries. Existing one-stage methods encode the entire language query as a single sentence embedding vector, \eg, taking the embedding from BERT or the hidden state from LSTM. This single vector representation is prone to overlooking the detailed descriptions in the query. To address this query modeling deficiency, we propose a recursive sub-query construction framework, which reasons between image and query for multiple rounds and reduces the referring ambiguity step by step. We show our new one-stage method obtains  $5.0\%, 4.5\%, 7.5\%, 12.8\%$ absolute improvements over the state-of-the-art one-stage approach on ReferItGame, RefCOCO, RefCOCO+, and RefCOCOg, respectively. In particular, superior performances on longer and more complex queries validates the effectiveness of our query modeling. Code is available at \url{https://github.com/zyang-ur/ReSC}.

\end{abstract}

\section{Introduction}
Visual grounding aims to ground a natural language query onto a region of the image. There are mainly two threads of works in visual grounding: the two-stage approach~\cite{wang2016learning,wang2019learning,plummerCITE2018,chen2017msrc,yu2017joint,yu2018mattnet} and one-stage approach~\cite{yang2019fast,chen2018real,sadhu2019zero}. Two-stage approaches first extract region proposals and then rank the proposals based on their similarities with the query. The recently proposed one-stage approach takes a different paradigm but soon becomes prevailing. The one-stage approach fuses visual-text features at image-level and directly predicts bounding boxes to ground the referred object. By densely sampling the possible object locations and reducing the redundant computation over region proposals, the one-stage methods~\cite{yang2019fast,chen2018real,sadhu2019zero} are simple, fast, and accurate. 

\begin{figure}[t]
\begin{minipage}{.4\textwidth}
  \centering
  \includegraphics[width=4.8cm]{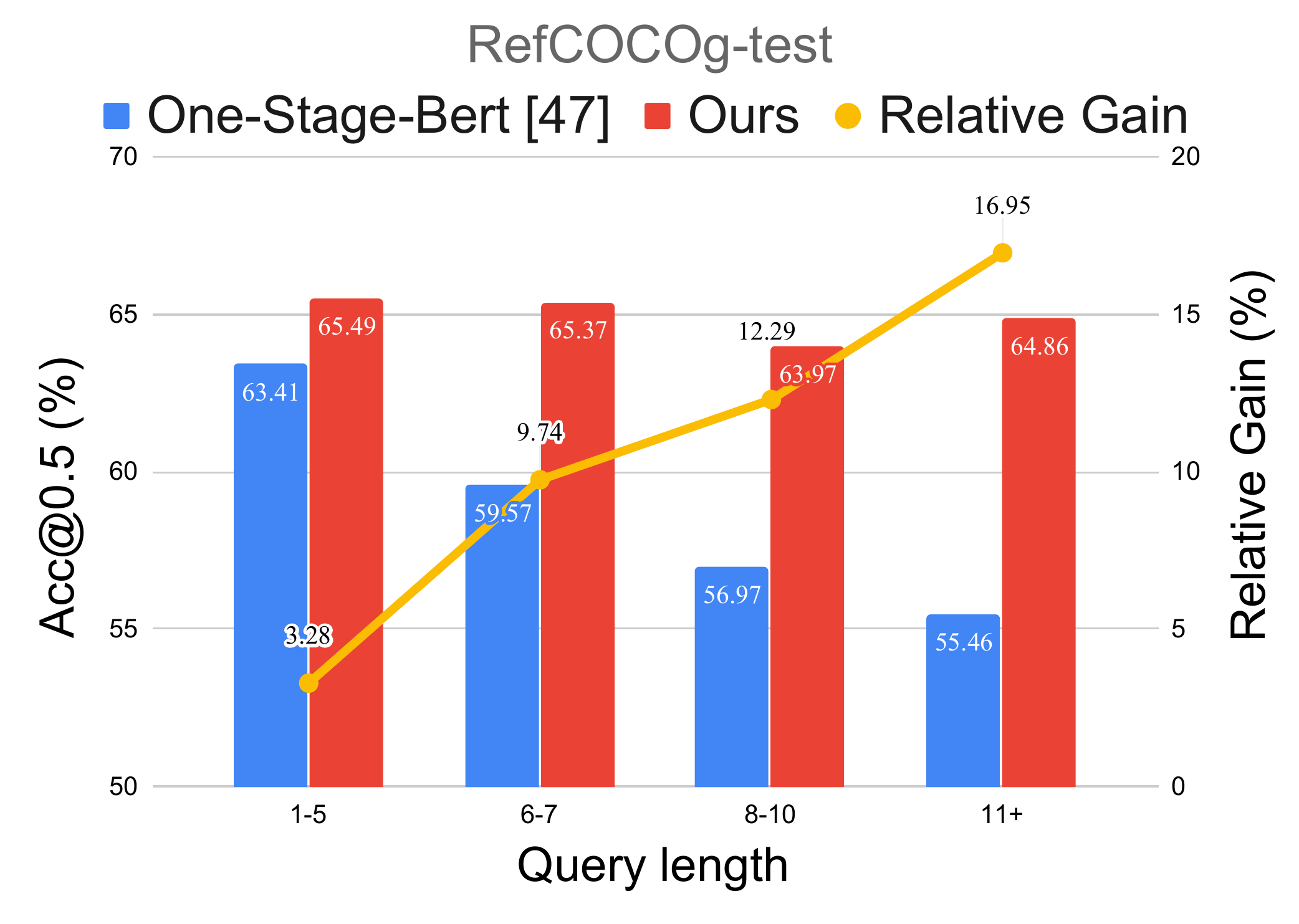}
  \caption{The accuracy of previous one-stage methods (blue column) decreases on longer queries.}
  \label{fig:chart}
\end{minipage}%
\qquad
\begin{minipage}{.55\textwidth}
  \centering
  \includegraphics[width=7cm]{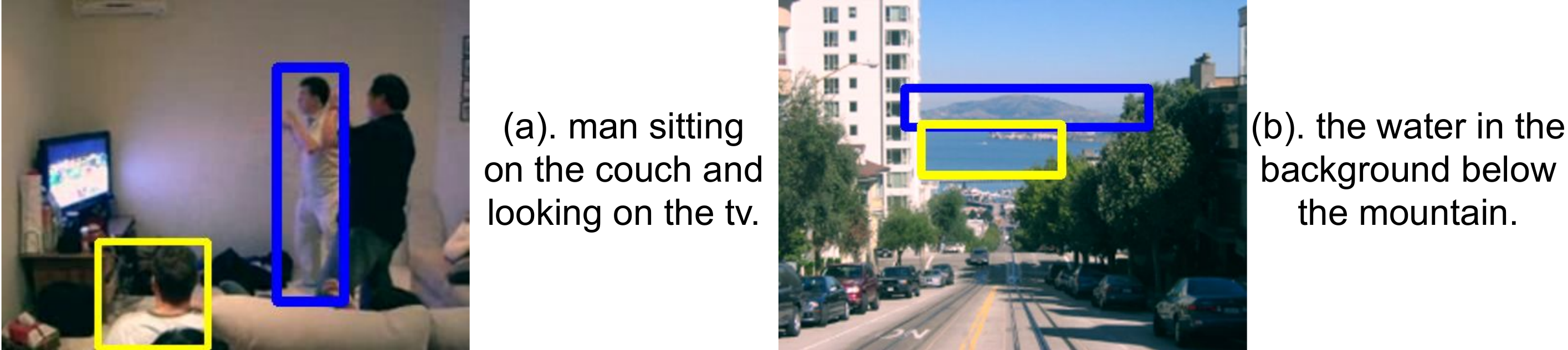}
  \caption{Previous one-stage methods' representative failure cases of (a) overlooking detailed descriptions, (b) misinterpreting the query by keywords. Blue/ yellow boxes are the predicted regions~\cite{yang2019fast}/ ground truths.}
  \label{fig:intro}
\end{minipage}
\end{figure}

In this paper, we improve the state-of-the-art one-stage methods by addressing their weaknesses in modeling long and complex queries. The overall advantage of our method is shown in Figure~\ref{fig:chart}. Compared to the current state-of-the-art one-stage method~\cite{yang2019fast}, whose performance \textit{drops dramatically on longer queries}, our approach achieves remarkably superior performance.

We analyze the limitations of current one-stage methods as follows. Existing one-stage methods~\cite{yang2019fast,chen2018real,sadhu2019zero} encode the entire query as a single embedding vector, such as directly adopting the first token's embedding ($\left[\textrm{CLS} \right]$) from BERT~\cite{devlin2018BERT,yang2019fast} or aggregating hidden states from LSTM~\cite{greff2016lstm,yang2019fast,chen2018real,sadhu2019zero}. The single vector is then concatenated at all spatial locations with visual features to obtain the fused features for grounding box prediction.
Modeling the entire language query as a single embedding vector tends to increase representation ambiguity, such as focusing on some words, yet overlooking other important ones. Such a problem potentially causes the loss of referring information, especially on those long and complex queries.\eat{The previous method\cite{yang2019fast} appears to either overlook certain detailed descriptions or misinterpret the keywords. \zyang{emphasize these are just two examples, not the solution motivation.}}
For example in Figure~\ref{fig:intro} (a), the model seems to overlook detailed descriptions such as ``sitting on the couch'' or ``looking tv,'' and grounds the wrong region with the head noun ``man.'' As for Figure~\ref{fig:intro} (b), the model appears to look into the wrong word ``mountain'' and grounds the target without full consideration of the modifier of ``water.'' Neglecting the query modeling problem, thus, causes the performance drop on long queries for the one-stage approach.

Several two-stage visual grounding works~\cite{wang2019neighbourhood,yang2019cross,yang2019dynamic,liu2019learning,yu2018mattnet,liu2019improving} have studied a similar query modeling problem. The main idea of these works is to link object regions with the parsed sub-queries to have a comprehensive understanding of the referring. Among them, MattNet~\cite{yu2018mattnet} parses the query into the subject, location, and relationship phrases, and links each phrase with the related object regions for matching score computing. NMTREE~\cite{liu2019learning} parses the query with a dependency tree parser~\cite{chen2014fast} and links each tree node with a visual region. DGA~\cite{yang2019dynamic} parses the query with text self-attention and links the text with regions via dynamic graph attention.\eat{ on an object visual graph. \zyang{do you want to add mattnet paper here? add one sentence to describe it}} Though elegant enough, these methods are designed intuitively for two-stage methods, requiring candidate region features to be extracted at the first stage. Since the main benefit of doing one-stage visual grounding is to avoid explicitly extracting candidate region features for the sake of computational cost, the query modeling in two-stage methods cannot be directly applied to the one-stage framework~\cite{yang2019fast,chen2018real,sadhu2019zero}.
Therefore, in this paper, to address the query modeling problem in a unified one-stage framework, we propose the recursive sub-query construction framework.

When presented with a referring problem such as Figure~\ref{fig:intro} (a), humans tend to solve it by reasoning back-and-forth between the image and query for multiple rounds and recursively reduce the referring ambiguity, \ie, the possible region that contains the referred object.
Inspired by this, we proposed to represent the intermediate understanding of the referring in each round as the \textit{text-conditional visual feature}, which starts as the image feature and updated after multiple rounds, ending up as the fused visual-text feature ready for box prediction. In each round, the model constructs a new sub-query as a group of words attended with scores to refine the text-conditional visual feature.\eat{\zyang{check the above sentence}} Gradually, with multiple rounds, our model reduces the referring ambiguity. Such a multi-round solution is in contrast to previous one-stage approaches that try to remember the entire query and ground the region in a single round. 

Our framework recursively constructs sub-queries to refine the grounding prediction. Each round faces with two core problems that facilitate recursive reasoning, namely 1) how to construct the sub-query; and 2) how to refine the text-conditional visual feature with the sub-query. We propose a sub-query learner and a sub-query modulation network to solve the above two problems, respectively.
They work alternately and recursively to reduce the referring ambiguity. Using the text-conditional visual features in the last round, a final output stage predicts bounding boxes to grounding the referred object.

We benchmark our framework on the ReferItGame~\cite{kazemzadeh2014referitgame}, RefCOCO~\cite{yu2016modeling}, RefCOCO+~\cite{yu2016modeling}, RefCOCOg~\cite{mao2016generation} datasets, with $5.0\%, 4.5\%, 7.5\%, 12.8\%$ absolute improvements over the state-of-the-art one-stage method~\cite{yang2019fast}. 
Meanwhile, our method runs fast at 38 FPS ($26ms$).
Moreover, the relative gain curve according to the query length changes in Figure~\ref{fig:chart} shows the effectiveness of our approach in solving the aforementioned query modeling problem.

Our main contributions are:
\begin{itemize} 
\item We improve one-stage visual grounding by addressing previous one-stage methods' limitations on grounding long and complex queries.
\item We propose a recursive sub-query construction framework that recursively reduces the referring ambiguity with different constructed sub-queries.
\item Our proposed method shows significantly improved results on multiple datasets and meanwhile maintains the real-time inference speed. Extensive experiments and ablations validate the effectiveness of our method.
\end{itemize}
\section{Related Work}
There exists two major categories of visual grounding methods: phrase localization~\cite{kazemzadeh2014referitgame,plummer2017flickr30k,wang2019learning} and referring expression comprehension~\cite{mao2016generation,yu2016modeling,yu2018mattnet,hu2016natural,li2017deep}.\eat{The query in phrase localization comes from a noun phrase in a full sentence and is generally of a short length. Referring expression comprehension usually contains longer queries that describe attributes of specific objects and their relationships.} Most previous visual grounding methods are composed of two stages. In the first stage, a number of region proposals are generated by an off-line module such as EdgeBox~\cite{zitnick2014edge}, selective search~\cite{uijlings2013selective} or pretrained detectors~\cite{liu2016ssd,ren2015faster,he2017mask}. In the second stage, each region is compared to the input query and outputs a similarity score. During inference, the region with the highest similarity score is output as the final prediction. Under the two-stage framework, various works explore different aspects to improve visual grounding, such as better exploiting attributes~\cite{liu2017referring,yu2018mattnet,liu2019improving}, object relationships~\cite{wang2019neighbourhood,yang2019cross,yang2019dynamic,liu2019learning}, phrase co-occurrences~\cite{chen2017query,dogan2019neural,bajaj2019g3raphground}, \etc

Recently, several works~\cite{yang2019fast,chen2018real,sadhu2019zero,liao2020real,luo2020multi} propose a different paradigm of one-stage visual grounding. The primary motivation is to solve the two limitations of two-stage methods, \ie, the performance cap caused by the sparsely sampled region proposals, and the slow inference speed caused by the region feature computation. Instead of explicitly extracting the features for all proposed regions, one-stage methods fuse the visual-text feature densely at all spatial locations, and directly predict bounding boxes to ground the target. Previous one-stage methods usually encode the query as a single language vector and concatenate the feature along the channel dimension of the visual feature. Despite the effectiveness of one-stage methods, modeling the language query as a single vector could lead to the loss of referring information, especially on long and complex queries. Though two-stage methods~\cite{wang2019neighbourhood,yang2019cross,yang2019dynamic,liu2019learning,yu2018mattnet,liu2019improving} had studied the similar problem of language query modeling, the explorations can not be directly applied to the one-stage approach given the intrinsic difference between two paradigms. 

Besides, an intuitive alternative is to model the query phrase by the attention mechanism. Lin~\etal~\cite{lin2017structured} propose to extract sentence embedding with self-attention. Modeling query with attention mechanism is also explored in various vision-language tasks~\cite{li2017person,yu2018mattnet}. In experiments, we observe that our proposed multi-round solution performs better than the simple query attention method (\cf ``Single/ Multi-head attention'' and ``Sub-query learner (ours)'' in Table~\ref{table:abla}).

\section{Approach}
\label{sec:method}

In this section, we will introduce our query modeling in a unified one-stage grounding framework. Previous one-stage grounding methods encode the language query as a $C_l$-dimension language feature and concatenate the text feature at all spatial locations with the visual feature $v\in R^{H\times W\times C_v}$. $H,W,C_v$ are the height, width, and dimension of the visual feature. The visual feature and the text feature are usually mapped to the same dimension $C$ before the concatenation. Extra convolutional layers then further refine the fused feature $f\in R^{H\times W\times 2C}$ and predict bounding boxes at each spatial location $H\times W$ to ground the target.
Such single-round query modeling tends to overlook important query details and lead to incorrect predictions\eat{failures on challenging samples where detailed query analyses are necessary}. The problems become increasingly severe on longer and more complex queries, as shown quantitatively in Figure~\ref{fig:chart}.

\begin{figure*}[t]
\centering
\includegraphics[width=12cm]{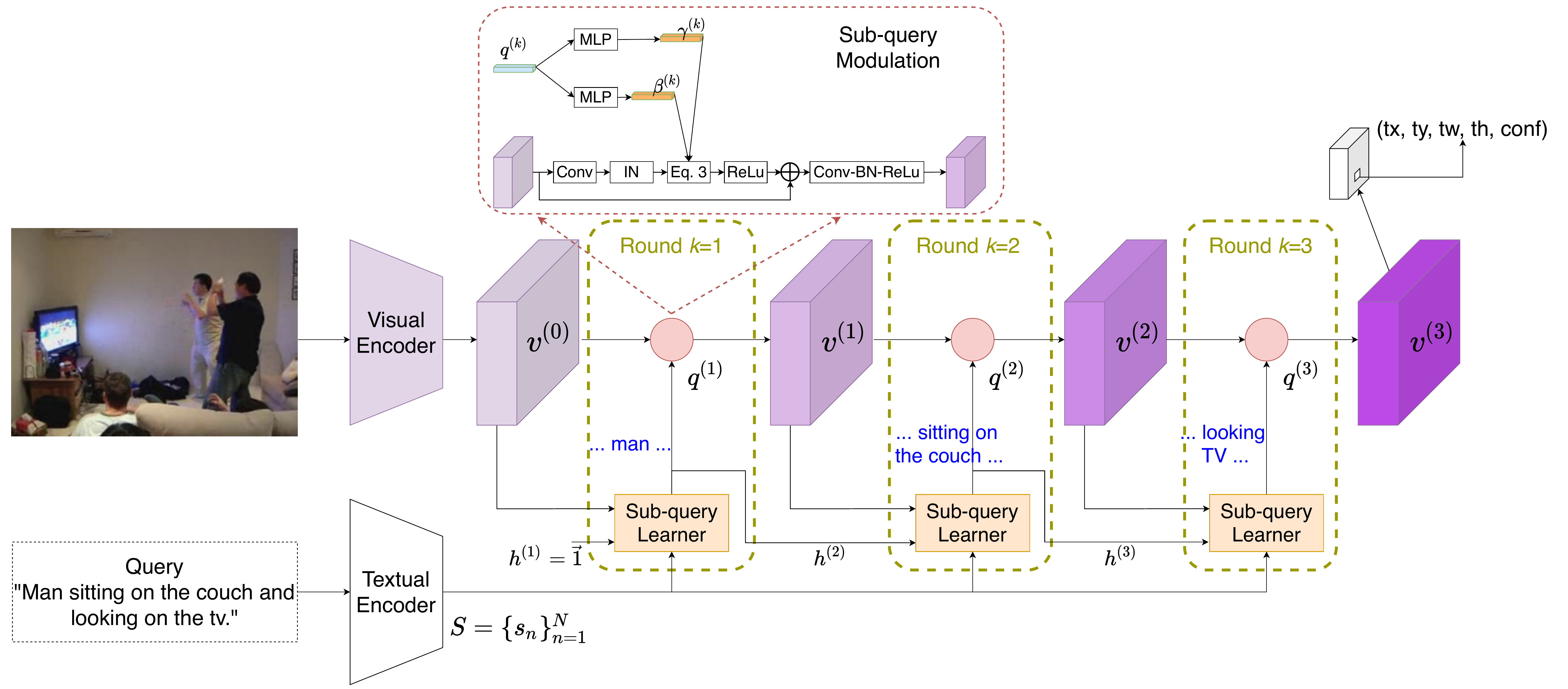}
    \caption{The architecture for the recursive sub-query construction framework. In each round, the framework constructs a new sub-query to refine the text-conditional visual feature $v^{(k)}$ (shown in purple). $q^{(k)}$ is the feature for the constructed sub-query.
    }
\label{fig:arch}
\end{figure*}

To address this problem in a unified one-stage grounding system, we propose a recursive sub-query construction framework that step by step refines the visual-text feature $v^{(k)}$ to get better prediction. 
As shown in Figure \ref{fig:arch}, the initial feature $v^{(0)}$ is the image feature $v\in R^{H\times W\times C}$ encoded by the visual encoder~\cite{redmon2018yolov3}. In each round $k$, the framework constructs a new sub-query as a group of words attended with score vector $\vec{\alpha}^{(k)}$, and obtains the sub-query embedding $q^{(k)}$ to refine the visual-text feature. The framework ends up with the refined feature $v^{(K)}$ after $K$ rounds, and predicts bounding boxes on each spatial location of $v^{(K)}$ to ground the target. We name $v^{(k)}$ the text-conditional visual feature.

In each round, we address how to construct the sub-query and how to refine the feature $v^{(k)}$ with the sub-query embedding to better ground the target. For the first problem, we propose a sub-query learner in Section \ref{sec:parse}. The objective is to construct the sub-query that could best resolve the current referring ambiguity. We find it important to refer to the text-conditional visual feature $v^{(k)}$ in each round. 
For the second problem, we propose a sub-query modulation network that scales and shifts the feature $v^{(k)}$ with the sub-query feature. We introduce the details in Section~\ref{sec:modu}. 
The sub-query learner and the modulation network operate alternately for multiple rounds, and recursively reduce the referring ambiguity. The feature $v^{(K)}$ in the last round contains the full object referring information and is used for target box prediction.
\subsection{Sub-query Learner}
\label{sec:parse}
Our method addresses the visual grounding problem as a multi-round reasoning process. In each round, the sub-query learner refers to the text-conditional visual feature $v^{(k)}$ and constructs the sub-query that could gradually reduce the referring ambiguity.

Given a language query of $N$ words, the language encoder extracts the per-word query representation $S=\{s_n\}_{n=1}^N$, each with the dimension of $C$.
As shown in Figure~\ref{fig:arch}, the sub-query learner constructs a sub-query in each round $k$ as a group of words attended with score vector $\vec{\alpha}^{(k)}=\{\alpha_n^{(k)}\}_{n=1}^N$ of length $N$. 
Apart from the query word features $S$, we find it particularly important to reference the current text-conditional visual feature $v^{(k-1)}\in R^{H\times W\times C}$ in sub-query construction, and thus take the average-pooled feature $\Bar{v}^{(k-1)}$ of dimension $C$ as an additional input to the learner. Moreover, the history of the previous sub-queries could help to avoid overemphasizing certain keywords. Given the history of the previous sub-queries $\{\vec{\alpha}^{(i)}\}_{i=1}^{k-1}$, the history vector $\vec{h}^{(k)}$ represents the words that have been previously attended on and is calculates as
\begin{equation*}
    \vec{h}^{(k)} = \vec{1}- \min\left(\sum_{i=1}^{k-1}\vec{\alpha}^{(i)},\vec{1}\right),
\end{equation*}
where $\vec{1}$ is the all-ones vector. Both $\vec{h}^{(k)}$ and $\vec{\alpha}^{(i)}$ are $N$-Dimension vectors with values range from $0$ to $1$. The sub-query learner takes the query word feature $\{s_n\}_{n=1}^N$, the text-conditional visual feature $\Bar{v}^{(k-1)}$, and the history vector $\vec{h}^{(k)}=\{h_n^{(k)}\}_{n=1}^N$ to construct the sub-query for round $k$ by predicting score vector $\vec{\alpha}^{(k)}$:
\begin{equation}
\label{equ:attn}
    \alpha_n^{(k)} = \textit{softmax} \left[W_{a1}^{(k)} \tanh \left(W_{a0}^{(k)} h_n^{(k)}(\Bar{v}^{(k-1)} \odot s_n)+b_{a0}^{(k)}\right) +b_{a1}^{(k)}\right],
\end{equation}
\eat{{\color{red} delete repetition -- where $\Bar{v}^{(k-1)}$ is the averaged pooled visual feature $v^{(k-1)}$,}} where $\odot$ represents hadamard product, and $W_{a0},b_{a0},W_{a1},b_{a1}$ are learnable parameters. We compute the softmax over the $N$ attention scores.

To guide the multi-round reasoning, explicit regularization is imposed on the word attention scores. Intuitively, the constructed sub-queries at each round should focus on different elements of the query, and in the end, most words in the query should have been looked at. Therefore, we add two regularization terms:
\begin{equation}
\label{equ:reg}
    L_{\textit{div}}= \left\Vert A^TA \odot \left(\mathds{1} - I \right)\right\Vert_F^2,
    \quad
    L_{\textit{cover}}= \left\Vert \vec{1}- \min\left(\sum_{i=1}^{K}\vec{\alpha}^{(i)},\vec{1}\right) \right\Vert_1,
\end{equation}
where matrix $A$ is the predicted attention score matrix $A=\{\alpha_n^{(k)}\}_{n,k=1,1}^{N,K}$, $\mathds{1}$ is the matrix of ones and $I$ is an identity matrix. $L_{\textit{div}}$ avoids any words being focused on in more than one round and thus enforces the diversity. $L_{\textit{cover}}$ helps the model looks at all words in the query and thus improves the coverage.

The adopted technique of attention-based sub-query learning is related to previous compositional reasoning studies~\cite{hudson2018compositional,yang2019dynamic}. The major difference is that our method refers to the text-conditional visual feature $v^{(k)}$ to recursively construct the sub-query in each round. 
In contrast, the sub-query learning in previous studies~\cite{hudson2018compositional,yang2019dynamic} is purely based on the word feature $\{s_n\}_{n=1}^N$, and generates all sub-queries in prior to visual-text fusion. 

\subsection{Sub-query Modulation}
\label{sec:modu}

In each round, the sub-query learner constructs a sub-query as a group of words attended by a score vector $\vec{\alpha}^{(k)}$, and generates the sub-query feature $q^{(k)}$ as
\begin{equation*}
    q^{(k)} = \sum_{n=1}^{N}\alpha_n^{(k)} s_n .
\end{equation*}
The goal for the sub-query modulation is to refine the text-conditional visual feature $v^{(k-1)}$ with the new sub-query feature $q^{(k)}$, such that the refined feature $v^{(k)}$ performs better in grounding box prediction.

Inspired by conditional normalization on image-level tasks~\cite{de2017modulating,dumoulin2016learned,perez2018film}, we encode the sub-query representation $q^{(k)}$ to modulate the previous visual-text representation $v^{(k-1)}$ by scaling and shifting. To be specific, $q^{(k)}$ is projected into a scaling vectors $\gamma^{(k)}$ and a shifting vector $\beta^{(k)}$ with two MLPs respectively:
\begin{equation*}
    \gamma^{(k)} = \tanh{ \left(W_{\gamma}^{(k)}q^{(k)}+b_{\gamma}^{(k)}\right) },
    \quad
    \beta^{(k)} = \tanh{ \left(W_{\beta}^{(k)}q^{(k)}+b_{\beta}^{(k)}\right) }.
\end{equation*}
The text-conditional visual feature $v^{(k)}$ is then refined from $v^{(k-1)}$ with the two modulation vectors and extra learnable parameters:
\begin{equation}
\label{equ:modu}
    v^{(k)}(i,j) = f_2\left\{\textit{ReLU} \left[ f_1(v^{(k-1)}(i,j)) \odot \gamma^{(k)} + \beta^{(k)} \right]+v^{(k-1)}(i,j) \right\},
\end{equation}
where $(i,j)$ are the spatial coordinates, $f_1,f_2$ are learnable mapping layers as shown in Figure~\ref{fig:arch}. $f_1$ consists of a $1\times1$ convolution followed by an instance normalization layer. $f_2$ consists of a $3\times3$ convolution followed by a batch normalization layer and ReLU activation. The grounding module takes the text-conditional visual feature in the final round $v^{(K)}$ as input, and predicts bounding boxes to ground the referred object.
With the extra referring information in each sub-query $q^{(k)}$, we expect the modulation in each round to strength the feature of the referred object, and meanwhile suppress the ones for distracting objects and the background.

Our proposed sub-query modulation in Equation~\ref{equ:modu} has shared modulation vectors $\gamma^{(k)},\beta^{(k)}$ over all spatial locations $(i,j)$. One intuitive alternative is to predict different modulation vectors $\gamma^{(k)}(i,j), \beta^{(k)}(i,j)$ for each spatial location. This can be done by constructing sub-queries $\vec{\alpha}^{(k)}(i,j)$ for each location with the corresponding text-conditional visual feature $v^{(k-1)}(i,j)$. Despite using different modulation parameters at different spatial locations seems more intuitive, we show that the modulation along the channel dimension achieves the same objective and meanwhile is computationally efficient (\cf ``Spatial-independent sub-query'' and ``Sub-query learner (ours)'' in Table~\ref{table:abla}).

\subsection{Framework Details}
\label{sec:out}

\noindent\textbf{Visual and text feature encoder.} We resize the input image to $3\times256\times256$ and use Darknet-53~\cite{redmon2018yolov3} pretrained on COCO object detection~\cite{lin2014microsoft} as the visual encoder. We adopt the visual feature from the 102-th convolutional layer that has a dimension $32\times32\times256$. We map the raw visual feature into the visual input $v^{(0)}$ with a $1\times1$ convolutional layer with batch normalization and ReLU. We set the shared dimension $C=512$.

We encode the each word in the query as a $768D$ vector with the uncased base version of BERT~\cite{devlin2018BERT,Wolf2019HuggingFacesTS}. We sum the representations for each word in the last four layers and map the features with two fully connected layers to obtain the representation $S=\{s_n\}_{n=1}^N$. $N$ is the number of query words and does not include special tokens such as $\left[\textrm{CLS} \right]$, $\left[\textrm{SEP} \right]$ and $\left[\textrm{PAD} \right]$.

\noindent\textbf{Grounding module.} The grounding module takes the visual-text feature $v^{(K)}$ as input and generates object prediction at each spatial location to ground the target. We use the same two $1\times 1$ convolutional layers as in One-Stage-BERT~\cite{yang2019fast} for box prediction. There are $32\times32=1024$ spatial locations, and we predict $9$ anchor boxes at each location. We follow the anchor selection steps in a previous one-stage grounding method~\cite{yang2019fast} with the same anchor boxes used. For each one of the $1024\times9=9216$ anchor boxes, we predict the relative offset and confidence score. A cross-entropy loss between the softmax over all the 9216 boxes and the one-hot target center vector, a regression loss of the relative location and size offset, and the regularization in Equation~\ref{equ:reg} are used to train the model. We use the same classification and regression losses as in One-Stage-BERT~\cite{yang2019fast}. 
\section{Experiments}

\subsection{Datasets}
\noindent{\bf RefCOCO/ RefCOCO+/ RefCOCOg.}  RefCOCO~\cite{yu2016modeling}, RefCOCO+~\cite{yu2016modeling}, and RefCOCOg~\cite{mao2016generation} are three visual grounding datasets with images and referred objects selected from MSCOCO~\cite{lin2014microsoft}. The referred objects are selected from the MSCOCO object detection annotations and belong to 80 object classes. RefCOCO~\cite{yu2016modeling} has 19,994 images with 142,210 referring expressions for 50,000 object instances. RefCOCO+ has 19,992 images with 141,564 referring expressions for 49,856 object instances. RefCOCOg has 25,799 images with 95,010 referring expressions for 49,822 object instances. On RefCOCO and RefCOCO+, we follow the split~\cite{yu2016modeling} of train/ validation/ testA/ testB that has 120,624/ 10,834/ 5,657/ 5,095 expressions for RefCOCO and 120,191/ 10,758/ 5,726/ 4,889 expressions for RefCOCO+, respectively. Images in ``testA'' are of multiple people and the ones in ``testB'' contain all other objects. The queries in RefCOCO+ contains no absolute location words, such as ``on the right'' that describes the object's location in the image. On RefCOCOg, we experiment with the splits of RefCOCOg-google\cite{mao2016generation} and RefCOCOg-umd~\cite{nagaraja2016modeling}, and refer to the splits as the val-g, val-u, and test-u in Table~\ref{table:main}. The queries in RefCOCOg are generally longer than those in RefCOCO and RefCOCO+: the average lengths are 3.61, 3.53, 8.43, respectively, for RefCOCO, RefCOCO+, RefCOCOg.

\noindent{\bf ReferItGame.} The ReferItGame dataset~\cite{kazemzadeh2014referitgame} has 20,000 images from SAIAPR-12~\cite{escalante2010segmented}. We follow a cleaned version of split~\cite{hu2016natural,rohrbach2016grounding,chen2017query}, which has 54,127, 5,842, and 60,103 referring expressions in train, validation, and test set, respectively. 

\noindent{\bf Flickr30K Entities.} Flickr30K Entities~\cite{plummer2017flickr30k} has 31,783 images with 427K referred entities. We follow the same split in previous works~\cite{plummer2017flickr30k,plummerCITE2018,wang2019learning}. We note that the queries in Flickr30K Entities are mostly short noun phrases and do not well reflect the difficulty of comprehensive phrase understanding. We still benchmark our method on Flickr30K Entities and compare it with other baselines for experiments completeness.
\subsection{Implementation Details}
\label{sec:detail}
\noindent \textbf{Training.}
Following the standard setting~\cite{redmon2018yolov3,yang2019fast}, we keep the aspect ratio of the input image and resize the long edge to 256. We then pad the resized image to a size of $256\times256$ with the mean pixel value. We follow the data augmentation in previous one-stage studies~\cite{redmon2018yolov3,yang2019fast}. The RMSProp~\cite{tieleman2012lecture} optimizer, with an initial learning rate of $10^{-4}$ is used to train the model with a batch size of $8$. The learning rate decreases by half every $10$ epochs for a total of $100$ epochs. We set the weight for $L_{div}, L_{cov}$ as $1$. We select $K=3$ as the default number of rounds and defer the related ablation studies to supplementary materials. 

\noindent \textbf{Evaluation.}
We follow the same $Acc@0.5$ evaluation protocol in prior works~\cite{plummer2017flickr30k,rohrbach2016grounding}. Given a language query, this metric is to consider the predicted region correct if its IoU is at least 0.5 with the ground truth bounding box. 

\subsection{Quantitative Results}
\begin{table}[t]\fontsize{6}{7}\selectfont
\centering
\caption{The performance comparisons (Acc@0.5\%) on RefCOCO, RefCOCO+, RefCOCOg (upper table), and ReferItGame, Flickr30K Entities (lower table). We highlight the best one-stage performance with \textbf{bold} and the best two-stage performance with \underline{underline}.
The \textit{COCO-trained detector} generates ideal proposals only for images in COCO. This leads to the two-stage methods' good performance on RefCOCO, RefCOCO+, RefCOCOg, as well as the accuracy drop on other datasets (lower table).
}
\begin{tabular}{ l | l | c c c | c c c | c c c | c}
    \hline
    \multirow{2}{*}{Method} & \multirow{2}{*}{Feature} &
    \multicolumn{3}{c|}{RefCOCO} & \multicolumn{3}{c|}{RefCOCO+} & \multicolumn{3}{c|}{RefCOCOg} & Time\\
     & & val & testA & testB & val & testA & testB & val-g & val-u & test-u & (ms)\\
    \hline
    \multicolumn{8}{l}{\textit{Two-stage Methods}}\\
    \hline
    MMI~\cite{mao2016generation} & {VGG16-Imagenet} & - & 64.90 & 54.51 & - & 54.03 & 42.81 & 45.85 & - & - & -\\
    Neg Bag~\cite{nagaraja2016modeling} & {VGG16-Imagenet} & - & 58.60 & 56.40 & - & - & - & - & - & 49.50 & -\\
    CMN~\cite{hu2017modeling} & {VGG16-COCO} & - & 71.03 & 65.77 & - & 54.32 & 47.76 & 57.47 & - & - & -\\
    ParallelAttn~\cite{zhuang2018parallel} & {VGG16-Imagenet} & - & 75.31 & 65.52 & - & 61.34 & 50.86 & 58.03 & - & - & - \\
    VC~\cite{zhang2018grounding} & {VGG16-COCO} & - & 73.33 & 67.44 & - & 58.40 & 53.18 & \underline{62.30} & - & - & - \\
    LGRAN~\cite{wang2019neighbourhood} & {VGG16-Imagenet} & - & 76.6 & 66.4 & - & 64.0 & 53.4 & 61.78 & - & - & - \\
    SLR~\cite{yu2017joint} & {Res101-COCO} & 69.48 & 73.71 & 64.96 & 55.71 & 60.74 & 48.80 & - & 60.21 & 59.63 & - \\
    MAttNet~\cite{yu2018mattnet} & {Res101-COCO} & \underline{76.40} & \underline{80.43} & \underline{69.28} & \underline{64.93} & \underline{70.26} & \underline{56.00} & - & \underline{66.67} & \underline{67.01} & 320 \\
    DGA~\cite{yang2019dynamic} & {Res101-COCO} & - & 78.42 & 65.53 & - & 69.07 & 51.99 & - & - & 63.28 & 341 \\
    \hline
    \hline
    \multicolumn{8}{l}{\textit{One-stage Methods}}\\
    \hline
    SSG~\cite{chen2018real} & {Darknet53-COCO} & - & 76.51 & 67.50 & - & 62.14 & 49.27 & 47.47 & 58.80 & - & 25 \\
    {One-Stage-BERT~\cite{yang2019fast}} & {Darknet53-COCO} & 72.05 & 74.81 & 67.59 & 55.72 & 60.37 & 48.54 & 48.14 & 59.03 & 58.70 & 23 \\
    {One-Stage-BERT*} & {Darknet53-COCO} & 72.54 & 74.35 & 68.50 & 56.81 & 60.23 & 49.60 & 56.12 & 61.33 & 60.36 & 23 \\
    Ours-Base & {Darknet53-COCO} & 76.59 & 78.22 & \textbf{73.25} & 63.23 & 66.64 & 55.53 & \eat{59.86}60.96 & 64.87 & 64.87 & 26 \\
    Ours-Large & {Darknet53-COCO} & \textbf{77.63} & \textbf{80.45} & 72.30 & \textbf{63.59} & \textbf{68.36} & \textbf{56.81} & \textbf{63.12} & \textbf{67.30} & \textbf{67.20} & 36 \\
    \hline
\end{tabular}
\newline
\newline

\begin{tabular}{ l | l | c | c | c}
    \hline
    \multirow{2}{*}{Method} & \multirow{2}{*}{Feature} & ReferItGame &
    Flickr30K Entities & Time\\
     & & test & test & (ms)\\
    \hline
    \multicolumn{4}{l}{\textit{Two-stage Methods}}\\
    \hline
    CMN~\cite{hu2017modeling} & VGG16-COCO & 28.33 & - & -\\
    VC~\cite{zhang2018grounding} & VGG16-COCO & 31.13 & - & - \\
    MAttNet~\cite{yu2018mattnet} & Res101-COCO & 29.04 & - & 320 \\
    Similarity Net~\cite{wang2019learning} & Res101-COCO & 34.54 & 60.89 & 184\\
    CITE~\cite{plummerCITE2018} & Res101-COCO & \underline{35.07} & \underline{61.33} & 196\\
    \hline
    \hline
    \multicolumn{4}{l}{\textit{One-stage Methods}}\\
    \hline
    SSG~\cite{chen2018real} & Darknet53-COCO & 54.24 & - & 25 \\
    ZSGNet~\cite{sadhu2019zero} & Res50-FPN & 58.63 & 63.39 & - \\
    One-Stage-BERT~\cite{yang2019fast} & Darknet53-COCO & 59.30 & 68.69 & 23 \\
    One-Stage-BERT* & Darknet53-COCO & 60.67 & 68.71 & 23 \\
    Ours-Base & Darknet53-COCO & 64.33 & 69.04 & 26 \\
    Ours-Large & Darknet53-COCO & \textbf{64.60} & \textbf{69.28} & 36 \\
    \hline
\end{tabular}
\label{table:main}
\end{table}
\noindent \textbf{Experiment settings.}
Table~\ref{table:main} reports visual grounding results on RefCOCO, RefCOCO+, RefCOCOg (the upper table), and ReferItGame, Flickr30K Entities (the lower table). The \textit{top part} of each table contain results of the state-of-the-art two-stage visual grounding methods~\cite{mao2016generation,nagaraja2016modeling,hu2017modeling,yu2017joint,yu2018mattnet,liu2019improving,zhuang2018parallel,zhang2018grounding,wang2019neighbourhood,yang2019dynamic,wang2019learning,plummerCITE2018}. The ``\textit{Feature}'' column lists the backbone and pretrained dataset used for proposal feature extraction. COCO-trained Faster-RCNN~\cite{ren2015faster} detector is used for region proposal generation for the experiments on RefCOCO~\cite{yu2016modeling}, RefCOCO+~\cite{yu2016modeling} and RefCOCOg~\cite{mao2016generation}. We quote the two-stage methods' results on ReferItGame and Flickr30K Entities reported by SSG~\cite{chen2018real} and One-Stage-BERT~\cite{yang2019fast} where Edgebox~\cite{zitnick2014edge} is used for proposal generation. 

The \textit{bottom part} of Table~\ref{table:main} compares the performance of our method to other state-of-the-art one-stage methods~\cite{chen2018real,sadhu2019zero,yang2019fast}. The ``\textit{Feature}'' column shows the adopted visual backbone and its pretrained dataset, if any. For a fair comparison, we modify One-Stage-BERT~\cite{yang2019fast} to have the exact same training details as ours, and observe a small accuracy improvement by the modification.\eat{\zyang{why use ``reimplement''? they have the code released link,right? not sure why we need to make the changes to have one stage-BERT*. you should write this part more clear.}} Specifically, we 1). encode the query as the averaged BERT word embedding instead of the BERT sentence embedding at the first token's position ($\left[\textrm{CLS} \right]$), 2). remove the feature pyramid network, and 3). follow the implementation details in Section~\ref{sec:detail}. We refer to the modified version ``\textit{One-Stage-BERT*}.'' Other than the state-of-the-art, we design and compare to additional alternatives to our methods such as ``\textit{single/ multi-head attention query modeling},'' ``\textit{per-word sub-query},'' \etc, in Section~\ref{sec:ablation} and Table~\ref{table:abla}.

We obtain our main results by the method described in Section~\ref{sec:method} and refer to it as ``\textit{Ours-Base}'' in Table~\ref{table:main}. Furthermore, we observe that a larger input image size of $512$ and a ConvLSTM~\cite{xingjian2015convolutional,choy20163d} grounding module increase the accuracy, but meanwhile slightly slow the inference speed. We refer to the corresponding model as ``\textit{Ours-Large}'' and analyze each modification in supplementary materials. 

\noindent \textbf{Visual grounding results.}
Our proposed method outperforms the state-of-the-art one-stage grounding methods~\cite{yang2019fast,chen2018real,sadhu2019zero} by over $5\%$ absolute accuracy on all experimented datasets.

The two-stage methods~\cite{yu2018mattnet,liu2019improving,zhuang2018parallel,zhang2018grounding,wang2019neighbourhood,yang2019dynamic} also show good performance on COCO-series datasets (\ie, RefCOCO, RefCOCO+, and RefCOCOg) by using the COCO-trained detector~\cite{ren2015faster}\eat{ to generate ideal region proposals on COCO images}. For example, we notice that MAttNet~\cite{yu2018mattnet} achieves comparable performance with our best model in RefCOCO+, though our best model obviously surpasses MAttNet on RefCOCO, RefCOCOg, and the testB of RefCOCO+. However, in the ReferItGame dataset, as listed in the lower part of Table~\ref{table:main}, MAttNet's accuracy drops dramatically. The findings of previous one-stage work ~\cite{yang2019fast,chen2018real} show that two-stage visual grounding methods rely highly on the region proposals quality.
Since RefCOCO/ RefCOCO+/ RefCOCOg are subsets of COCO and have shared images and objects, the COCO-trained detector generates nearly perfect region proposals on COCO-series datasets. When used in other datasets, e.g., ReferItGame and Flickr30K Entities datasets, their proposal quality and grounding accuracy drop, such as the MAttNet's degraded performance in ReferItGame. Nonetheless, our method performs stably across all datasets and, meanwhile being significantly faster.

\noindent \textbf{Inference time.}
The real-time inference speed is one major advantage of the one-stage visual grounding method. We conduct all the experiments on a single NVIDIA 1080TI GPU. We observe our method achieves a real-time inference speed of $26ms$. The method is more than $10$ times faster than typical two-stage methods such as the MattNet~\cite{yu2018mattnet} of $320ms$.

\subsection{Performance break-down studies}
\label{sec:break}
We show the effectiveness of our method in modeling long queries by breaking down the test set. We split the test set of ReferItGame~\cite{kazemzadeh2014referitgame}, RefCOCO~\cite{yu2016modeling}, RefCOCO+~\cite{yu2016modeling}, and RefCOCOg~\cite{mao2016generation} each into four sub-sets based on the query lengths (we combine the testA and testB for RefCOCO and RefCOCO+). Table~\ref{table:breaklen} compares our method to One-Stage-BERT~\cite{yang2019fast} on the generated sub-sets. We adopt ``Ours-Base'' instead of ``Ours-Large'' for comparison because the inference speed of ``Ours-Base'' is more comparable with One-Stage-BERT.
The first row shows the experimented dataset and the number of query words in each sub-set. The second row shows the portion of samples in each subset. We generate sub-sets that are roughly with the same size. The middle two rows compare the accuracy of our method to the state-of-the-art one-stage grounding method~\cite{yang2019fast}. The last row computes the relative gain obtained by our method as $\left(\textit{Ours}-\textit{Base}\right)/\textit{Base}$. We observe a larger relative gain of our method on longer queries. The relative gain is around $20\%$ on the longest query sub-set. The consistent increases in the relative gain on all experimented datasets suggest the effectiveness of our recursive sub-query construction framework in modeling and grounding long queries.

\begin{table}[t]\small
\centering
\caption{The performance break-down with query lengths. The first row shows the experimented dataset and the number of query words in each sub-set.}
\parbox{.45\linewidth}{
\centering
\begin{tabular}{ l c c c c  }
    \hline
    \textit{RefCOCO} & 1-2 & 3 & 4-5 & 6+\\
    \hline
    Percent (\%) & 36.22 & 23.87 & 25.60 & 14.30 \\
    One-Stage-BERT & 77.68 & 76.04 & 66.98 & 55.59 \\
    Ours-Base & 79.35 & 79.28 & 72.65 & 66.19 \\
    \hline
    \textbf{Relative Gain} & 2.15 & 4.26 & 8.46 & 19.07 \\
    \hline\end{tabular}}
\quad
\quad
\parbox{.45\linewidth}{
\centering
\begin{tabular}{ l c c c c  }
    \hline
    \textit{RefCOCO+} & 1-2 & 3 & 4-5 & 6+\\
    \hline
    Percent (\%) & 37.79 & 19.48 & 27.40 & 15.33 \\
    One-Stage-BERT & 66.59 & 55.42 & 47.40 & 39.03 \\
    Ours-Base & 71.08 & 60.01 & 56.24 & 49.35 \\
    \hline
    \textbf{Relative Gain} & 6.74 & 8.28 & 18.65 & 26.44 \\
    \hline
\end{tabular}}
\newline
\newline

\parbox{.45\linewidth}{
\centering
\begin{tabular}{ l c c c c  }
    \hline
    \textit{RefCOCOg} & 1-5 & 6-7 & 8-10 & 11+\\
    \hline
    Percent (\%) & 23.54 & 22.80 & 28.30 & 25.37 \\
    One-Stage-BERT & 63.41 & 59.57 & 56.97 & 55.46 \\
    Ours-Base & 65.49 & 65.37 & 63.97 & 64.86 \\
    \hline
    \textbf{Relative Gain} & 3.28 & 9.74 & 12.29 & 16.95 \\
    \hline\end{tabular}}
\label{table:breaklen}
\quad
\quad
\parbox{.45\linewidth}{
\centering
\begin{tabular}{ l c c c c  }
    \hline
    \textit{ReferItGame} & 1 & 2 & 3-4 & 5+\\
    \hline
    Percent (\%) & 25.78 & 16.76 & 31.53 & 25.93 \\
    One-Stage-BERT & 82.33 & 66.66 & 56.64 & 34.89 \\
    Ours-Base & 82.12 & 69.46 & 61.43 & 46.84 \\
    \hline
    \textbf{Relative Gain} & -0.26 & 4.20 & 8.46 & 34.25 \\
    \hline
\end{tabular}}
\end{table}

\subsection{Ablation studies}
\label{sec:ablation}
In this section, we conduct ablation studies to understand our method better. We perform the study on RefCOCOg-google~\cite{mao2016generation} as it has, on average,  longer queries than other datasets, which can better reflect the query modeling problem.

\noindent \textbf{Query modeling.}
Table~\ref{table:abla} shows the ablation studies on different query modeling choices. Specifically, we systematically study the following settings.
\begin{itemize}
    \item\textbf{Average vector.} 
    We average the BERT embedded word features $S=\{s_n\}_{n=1}^N$ to form a single $512D$ vector as the query representation.\eat{If the experimented framework has multiple rounds (\ie, $K\geq2$), the feature vector is duplicated as the input to each round.}
    \item\textbf{Per word sub-query.} We consider each word as a sub-query. \eat{The number of rounds equals the number of words $N$. }The per-word sub-query modeling is used by RMI~\cite{liu2017recurrent} for referring image segmentation.
    \item\textbf{Single-head attention.} A set of self-attention scores of size $N$ is learned from the word features $\{s_n\}_{n=1}^N$. We obtain a single query feature vector by weighted sum the BERT embedded word features. We use the same self-attention method as in Equation~\ref{equ:attn} to obtain the attention scores~\cite{lin2017structured}, expect only the text feature $s_n$ is used as the input.
    \eat{ The query feature is duplicated as the input to each round in a multi-round framework.}
    \item\textbf{Multi-head attention.} Self-attention scores of size $N \times K$ are learned from the word features. One unique sub-query feature vector is formed as the input to each round.
    \item\textbf{Spatial-independent sub-query.} We discuss one alternative to our approach in the end of Section~\ref{sec:modu}. We refer to it as ``Spatial-independent sub-query'' as shown in the last row of Table~\ref{table:abla}.
    \item\textbf{Sub-query learner (ours).} \eat{\zyang{not clear how this one is working?} }Instead of jointly predicting the sub-queries for all steps, our proposed sub-query learner, as introduced in Section~\ref{sec:parse}, recursively constructs the sub-query by referring to the current text-conditional visual feature $v^{(k)}$.
\end{itemize}

Our proposed sub-query learner boosts the baseline accuracy with no attention by $1.7\%$ (\cf ``Average vector'' and ``Sub-query learner (ours)''). The query attention without the visual contents shows limited improvements over the no attention baseline (\cf ``Average vector'' and ``Single/ Multi-head attention''). Instead, by referring to the text-conditional visual feature in each round, our proposed sub-query learner further improve the attention baseline by $1.5\%$ (\cf ``Single/ Multi-head attention'' and ``Sub-query learner (ours)''). This shows the importance of recursive sub-query construction. Furthermore, ``spatial-independent sub-query'' constructs the sub-query independently at each spatial location. This alternative leads to extra computation while is not more accurate.

\begin{table}[t]\small
\centering
\caption{Ablation studies on query modeling. The sub-query modulation introduced in Section~\ref{sec:modu} is used for fusion.}
\begin{tabular}{ l c }
    \hline
    Query Modeling & Acc@0.5\\
    \hline
    Average vector & 59.24 \\
    Per word sub-query & 58.36 \\
    Single-head attention & 59.43 \\
    Multi-head attention & 59.25 \\
    Spatial-independent sub-query & 60.81 \\
    Sub-query learner (ours) & \textbf{60.96} \\
    \hline
\end{tabular}
\label{table:abla}
\end{table}

\noindent \textbf{Sub-query modulation.}
We compare with the ``Concat-Conv'' fusion used in One-Stage-BERT~\cite{yang2019fast}. To be specific, the query feature is duplicated spatially and is concatenated with the visual and spatial features to form a $512+512+8=1032D$ feature. One $1\times 1$ and one $3\times 3$ convolution layers then generate a $512D$ fused feature. In contrast, the sub-query modulation introduced in Section~\ref{sec:modu} converts the sub-query feature into scaling and shifting parameters to refine the text-conditional visual feature $v^{(k)}$.
Our proposed sub-query modulation improves the accuracy by $1.8\%$ with the similar amount of fusion parameters (\cf ``Concat-Conv'': $59.20\%$ and ``Ours'': $60.96\%$).

\begin{figure*}[t]
\centering
\includegraphics[width=12cm]{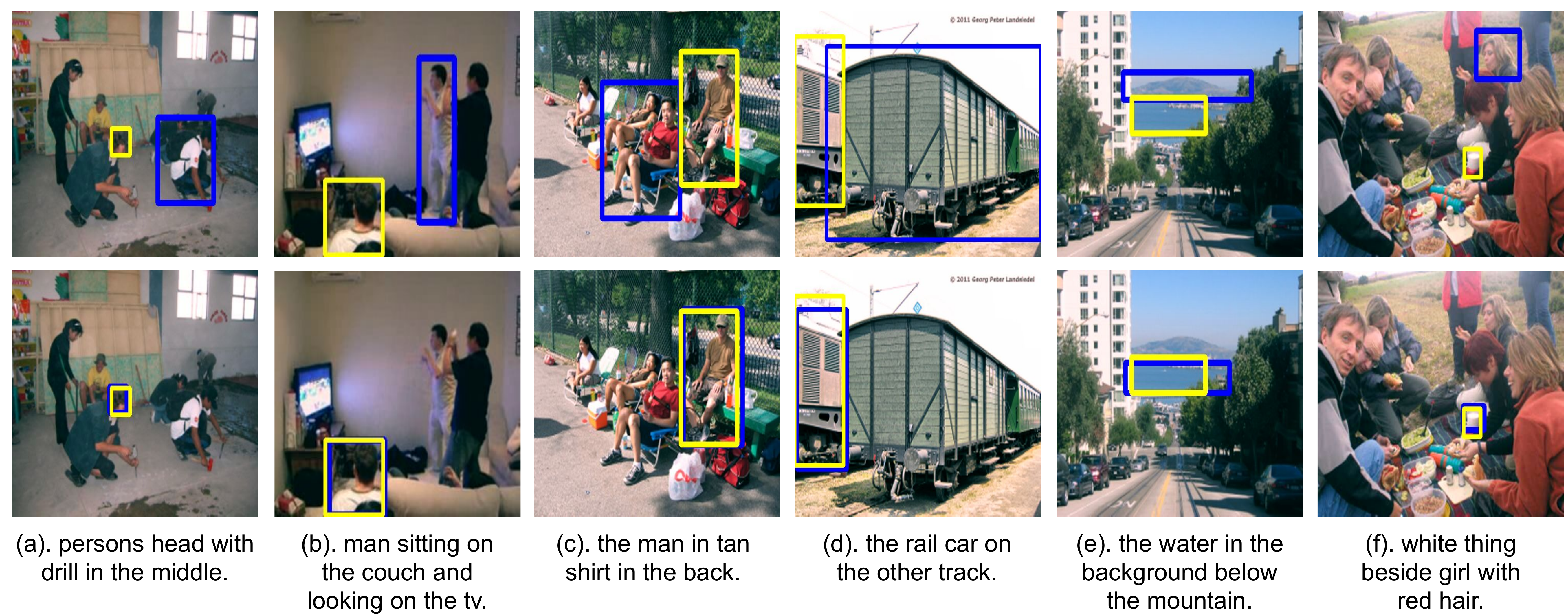}
    \caption{Failure cases of One-Stage-BERT~\cite{yang2019fast} (top row) that can be corrected by our method (bottom row). Blue/ yellow boxes are the predicted regions/ ground truths. Constructed sub-queries and per-round visualizations are in Figure~\ref{fig:reason}.
    }
\label{fig:comp}
\end{figure*}
\begin{figure*}[t]
\centering
\includegraphics[width=11.2cm]{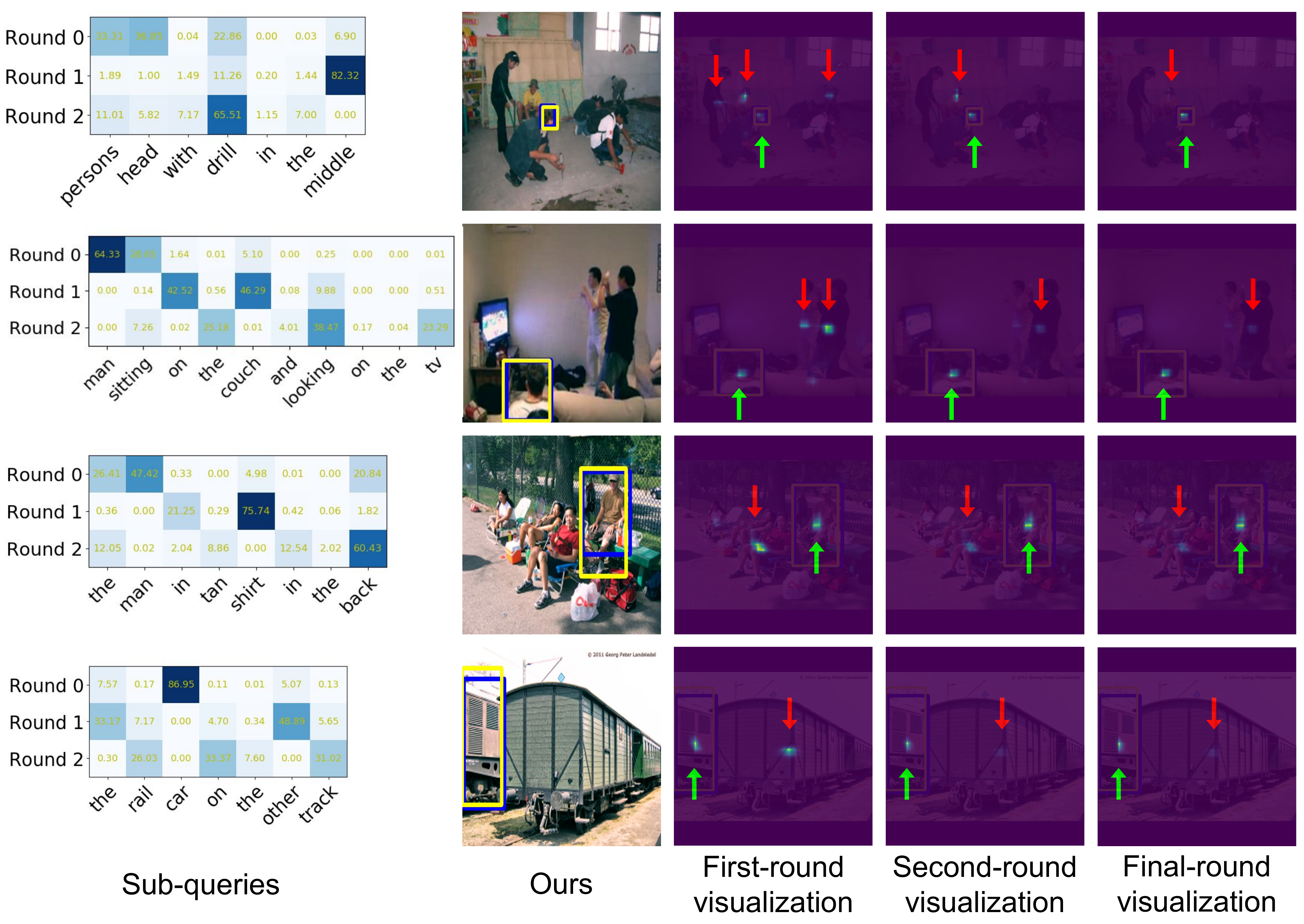}
    \caption{Visualization of the constructed sub-queries and the intermediate text-conditional visual feature at each round. The green up arrow and the red down arrow point to the target and the major distracting object on heatmaps, respectively. Bested viewed in color. More examples and detailed analyses are in supplementary materials.
    }
\label{fig:reason}
\end{figure*}
\subsection{Qualitative results}
Figure~\ref{fig:comp} shows the failure cases made by a previous one-stage method~\cite{yang2019fast} that can be correctly predicted by our method. We observe that previous methods appear to fail because of neglecting those detailed descriptions or modifiers (\eg, Figures~\ref{fig:comp} (a)-(d)), or attending on the wrong keyword (\eg, Figures~\ref{fig:comp} (e),(f)). In contrast, our method corrects these errors by better modeling the query. 

More importantly, in order to understand what happened inside our model and explain why it works, we show the visualization of intermediate results of our model in Figure~\ref{fig:reason}. The left column in Figure~\ref{fig:reason} shows the constructed sub-query in each round. The right three columns visualize the intermediate text-conditional visual feature $v^{(k)}$. For visualization, we adopt an extra output head over the feature $v^{(k)}$ in all steps and obtain the confidence score heatmaps. The confidence score indicates the probability of the object center. Therefore, the heatmap contains the peaks of object centers instead of the object contours. We highlight the referred object and the major distracting object with the green up arrow and red down arrow, respectively. 
We note that the intermediate prediction is just for visualization purpose, and is not in our proposed framework.

We observe that the model tends to focus on all head nouns in the first round, \eg, ``man'', ``head'', ``car'',\etc, because such keywords are the most informative sub-query when given a raw image. 
Then, in the next few rounds, our method can refine the intermediate text-conditional visual representation and reduce the referring ambiguity. For example, in the first row of Figure~\ref{fig:reason}, the model first focuses on the head-noun ``head''. Our model refines its prediction by the constructed sub-queries ``in the middle'' and ``with drill'' in the following rounds. Accordingly, from the heatmap visualization, we observe such a disambiguation process that the refined visual-text feature step by step generates more accurate and confident predictions. In the first round with the sub-query ``persons head,'' the model predicts four peaks in the heatmap, each centering at an appeared person. In the second round with sub-query ``in the middle,'' the model focuses on the two person in the middle and eliminates two distracting objects. In the final round with the sub-query ``with drill,'' the model successfully focuses on the referred person, and the heatmap values for all other distracting objects are greatly suppressed. We observe similar recursive disambiguation processes in other examples in Figure~\ref{fig:reason} and supplementary materials.

\section{Conclusions}
We have proposed a recursive sub-query construction framework to address the limitation of previous one-stage methods when understanding complex queries. We recursively construct sub-queries to refine the visual-text feature for grounding box prediction.
Extensive experiments and ablation studies have validated the high effectiveness of our method.  Our proposed framework significantly outperforms the state-of-the-art one-stage methods by over $5\%$ in  absolute accuracy on multiple datasets while still maintaining a real-time inference speed.  

\subsection*{Acknowledgment}
This work is supported in part by NSF awards IIS-1704337, IIS-1722847, and IIS-1813709, Twitch Fellowship, as well as our corporate sponsors.

\clearpage
%
%

\clearpage
\appendix
\section*{Appendix}
\section{Qualitative Results}
\label{sec:fig}
\begin{figure*}
\centering
\includegraphics[width=13cm]{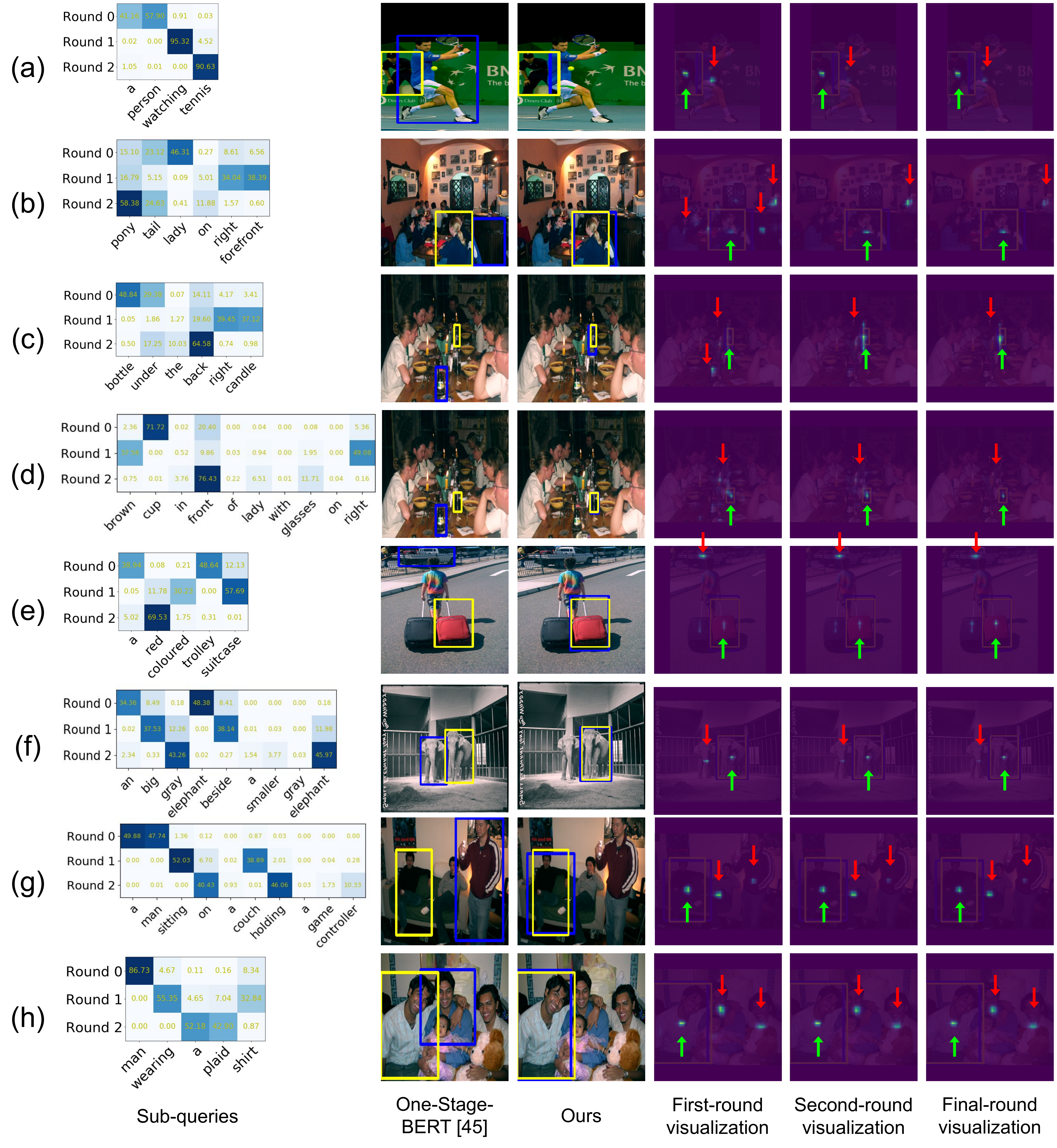}
    \caption{Visualizations of the constructed sub-queries and the intermediate text-conditional visual feature at each round. Blue/ yellow boxes are the predicted regions/ ground truths. The green up arrow and the red down arrow highlight the target and the major distracting objects on heatmaps, respectively.
    }
\label{fig:supply1}
\end{figure*}

\begin{figure*}[t]
\centering
\includegraphics[width=13cm]{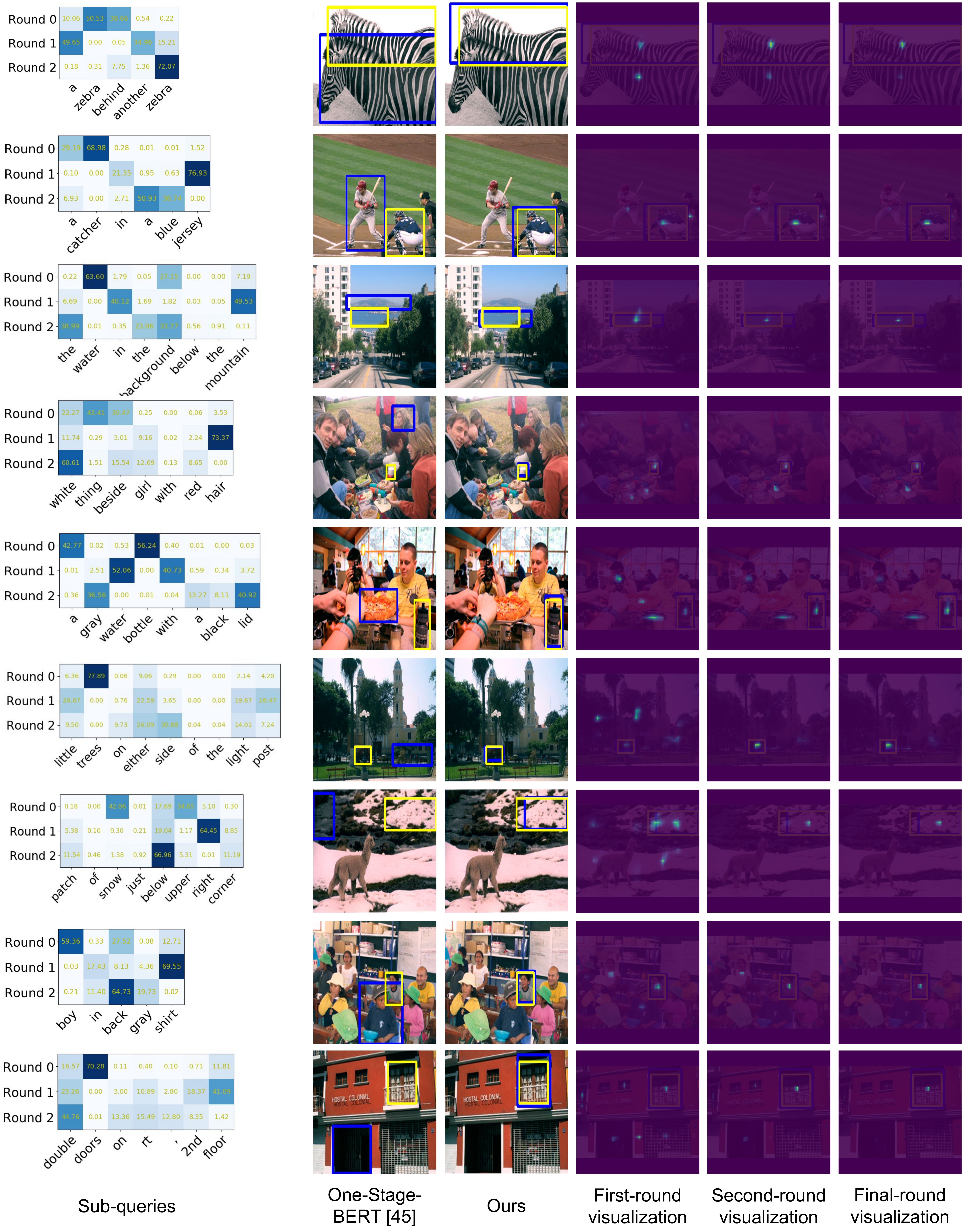}
    \caption{Additional qualitative results.
    }
\label{fig:supply2}
\end{figure*}

We present additional qualitative results in Figures~\ref{fig:supply1},~\ref{fig:supply2} to visualize the recursive disambiguation process. We highlight the following scenarios.

\begin{itemize}
    \item \textbf{Recursive disambiguation.} 
    The proposed recursive sub-query construction framework improves one-stage visual grounding by addressing the current limitations on grounding long queries. We observe a desired recursive disambiguation process that the text-conditional visual feature step by step generates more accurate and confident predictions.
    
    As an easy case, better modeling the modifiers of the head noun already corrects a portion of previous failures. For example, in Figure~\ref{fig:supply1} (a), the peak in the heatmap moves from the tennis player to the referred person in the back, after observing the modifier ``watching'' in the second round. On the contrary, previous one-stage methods tend to overemphasize the head nouns, without full consideration of the modifiers.
    Our proposed method also works well on more complex queries via recursive disambiguation, such as in the example ``persons head with drill in the middle'' from the main paper, and ``pony tail lady on the right forefront'' in Figure~\ref{fig:supply1} (b).
    \item \textbf{Challenging regions.}
    We observe that our method performs well on challenging examples such as Figures~\ref{fig:supply1} (c) and (d), where the referred target is tiny, the scene contains visually similar distracting objects, and the query includes complex attributes and relationship descriptions.
    \item \textbf{Attributes.}
    Figures~\ref{fig:supply1} (e) and (f) include examples that require the correct understanding of attributes such as color and size. For example, in the final round of Figure~\ref{fig:supply1} (e), the peak moves from the distracting object ``trolley'' to the referred red suitcase by observing the constructed sub-query ``red.''
    \item \textbf{Failure cases.}
    Figures~\ref{fig:supply1} (g) and (h) show failure cases of our model. The model either misses certain related objects such as the ``controller'' in Figures~\ref{fig:supply1} (g) or fails to understand some rarely appeared attributes such as ``plaid'' in Figures~\ref{fig:supply1} (h). Therefore, one or more distracting objects remain to have high heatmap responses in the final round, despite the model might still predict the correct bounding box.
\end{itemize}
We present additional qualitative results in Figure~\ref{fig:supply2}.

\section{Ablation Studies}
\label{sec:abla}

\begin{table}[t]\small
\parbox{.56\linewidth}{
\centering
\caption{Ablation studies on number of rounds.}
\begin{tabular}{ l c c}
    \hline
    \#Rounds \qquad & Acc@0.5 \qquad & Time(ms)\\
    \hline
    $K=1$ & 58.22 & 23\\
    $K=2$ & 59.31 & 25\\
    $K=3$ (Ours) & 60.96 & 26\\
    $K=4$ & 61.08 & 28\\
    $K=5$ & 60.80 & 30\\
    $K=6$ & 61.00 & 32\\
    \hline
\end{tabular}
\label{table:kabla}
}
\quad
\parbox{.4\linewidth}{
\centering
\caption{Ablation studies on the modifications in Ours-Large.}
\begin{tabular}{ l c c }
    \hline
    \quad  Method & Acc@0.5 & Time(ms)\\
    \hline
    \quad  Ours-Base & 60.96 & 26\\
    + ConvLSTM & 61.26 & 27 \\
    + Size 512 & 61.99 & 34\\
    + Both & 63.12 & 36 \\
    \hline
\end{tabular}
\label{table:fabla}}
\end{table}

\noindent \textbf{Number of rounds.} 
Table~\ref{table:kabla} shows the ablation studies on the different number of rounds $K$ on the RefCOCOg-google dataset. We observe that increasing the number of rounds does not lead to an increase in accuracy after a dataset-specific threshold (\eg, $K\geq3$ on RefCOCOg). Therefore, we select $K=3$ as the default value in our experiments for a balance between efficiency and accuracy. Although we report all results with $K=3$ in the main paper's Table 1 for clarity, we note that a different $K$ might slightly improve the accuracy or reduce the inference time on different datasets.

\noindent \textbf{Ours-Large.}
We observe several modules and settings that further improve the grounding accuracy, but meanwhile slightly slow the inference speed or increase the model complexity. Therefore, we list such modifications separately and refer to the corresponding framework ``Ours-Large.'' As shown in Table~\ref{table:fabla}, we observe an increase in accuracy with a larger input image size. Furthermore, we improve the accuracy by using all intermediate text-conditional visual feature with a ConvLSTM module. The ConvLSTM module takes feature $\{v^{(k)}\}_{k=1}^K$ as the input, and outputs the last hidden state for grounding box prediction.

\begin{table}[t]\fontsize{8}{9}\selectfont
\centering
\caption{Performance break-down with attributes.}
\parbox{.45\linewidth}{
\centering
\begin{tabular}{ l c c c | c  }
    \hline
    \textit{ReferItGame} & Color & Loc. & Size & All \\
    \hline
    Percent & 7.84 & 53.63 & 7.00 & 100.00 \\
    One-Stage-BERT & 43.07 & 50.98 & 53.83 & 59.30 \\
    Ours-Base & 50.52 & 58.71 & 61.83 & 64.33 \\
    \hline
    \textbf{Relative Gain} & 17.30 & 15.16 & 14.86 & 8.48 \\
    \hline
\end{tabular}}
\quad
\quad
\parbox{.45\linewidth}{
\centering
\begin{tabular}{ l c c c | c  }
    \hline
    \textit{RefCOCOg} & Color & Loc. & Size & All \\
    \hline
    Percent & 18.54 & 32.10 & 12.48 & 100.00 \\
    One-Stage-BERT & 55.06 & 56.46 & 57.85 & 58.70 \\
    Ours-Base & 62.92 & 66.90 & 66.53 & 64.87 \\
    \hline
    \textbf{Relative Gain} & 14.28 & 18.49 & 15.00 & 10.51 \\
    \hline\end{tabular}}
\label{table:breakcolor}
\end{table}

\noindent \textbf{Performance break-down with attributes.} 
The performance break-down study in Section 4.4 shows the effectiveness of our proposed method in modeling and grounding long queries. Other than the improvements on long queries, our method also shows advantages in modeling queries with attribute descriptions, such as color, location, or size. To validate the observation, we construct the attribute subsets from the test set of ReferItGame and RefCOCOg based on the contained attribute keywords, \eg, \textit{``white,'' ``black,'' ``red,'' ``blue,'' \etc,} for ``color;'' \textit{``right,'' ``left,'' ``front,'' ``middle,'' \etc,} for ``location;'' \textit{``big,'' ``little,'' ``small,'' ``tall,'' \etc,} for ``size.'' 

The first row of Table~\ref{table:breakcolor} shows the experimented dataset and the name of the subset. ``Color,'' ``location,'' and ``size'' indicate that the query in the subset contains at least one corresponding attribute keywords. ``All'' reports the performance on the entire dataset. The second row shows the portion of samples in each subset. The remaining rows indicate the grounding accuracy and the relative gain.
As shown in the last row, the relative gains on the attribute subsets are around $15\%$ and are higher than the relative gain on the entire dataset of around $10\%$ (\cf the middle three and the last column of Table~\ref{table:breakcolor}).

\end{document}